\documentclass[11pt,a4paper]{article}
\usepackage{authblk}
\usepackage[hyperref]{acl2020}
\usepackage{times}
\usepackage{latexsym}
\usepackage{numprint}
\usepackage{multirow}
\usepackage{soul}
\usepackage[table]{colortbl}

\usepackage{microtype}
\usepackage{todonotes}

\aclfinalcopy % Uncomment this line for the final submission
\title{A Multilingual Neural Machine Translation Model for Biomedical Data}

\author[*]{Alexandre B\'erard}
\author[**]{Zae Myung Kim}
\author[*]{Vassilina Nikoulina}
\author[**]{Eunjeong L. Park}
\author[*]{Matthias Gall\'e}
\affil[*]{Naver Labs Europe}
\affil[**]{Naver Papago}

\date{}

\begin{document}
\maketitle
\begin{abstract}
We release a multilingual neural machine translation model, which can be used to translate text in the biomedical domain.
The model can translate from 5 languages (French, German, Italian, Korean and Spanish) into English. It is trained with large amounts of generic and biomedical data, using domain tags. Our benchmarks show that it performs near state-of-the-art both on news (generic domain) and biomedical test sets, and that it outperforms the existing publicly released models.
We believe that this release will help the large-scale multilingual analysis of the digital content of the COVID-19 crisis and of its effects on society, economy, and healthcare policies.
We also release a test set of biomedical text for Korean-English.
It consists of 758 sentences from official guidelines and recent papers, all about COVID-19.

\end{abstract}

\section{Motivation}

The 2019--2020 coronavirus pandemic has disrupted lives, societies and economies across the globe.
Its classification as a \textit{pan}demic highlights its global impact, touching people of all languages.
Digital content of all types (social media, news articles, videos) have focused for many weeks predominantly on the sanitary crisis and its effects on infected people, their families, healthcare workers and the society and economy at large.
This calls not only for a large set of tools to help during the pandemic (as evidenced by the submissions to this workshop), but also for tools to help digest and analyze this data after it ends.
By analyzing the representation and reaction across countries with different guidelines or global trends, it might be possible to inform policies in prevention of and reaction to future epidemics. % AB: "span countries"? and weird formulation
Several institutions and groups have already started to take snapshots of the digital content shared during these weeks~\citep{lemonde2020,banda2020}.

However, because of its global scale, all this digital content is accessible in a variety of different languages, and most existing NLP tools remain English-centric~\citep{anastasopoulos2019should}.
In this paper we describe the release of a multilingual neural machine translation model (MNMT) that can be used to translate biomedical text.
The model is both multi-domain and multilingual, covering translation from French, German, Spanish, Italian and Korean to English.

Our contributions consist in the release of:
\begin{itemize}
    \item An MNMT model, and benchmark results on standard test sets;
    \item A new biomedical Korean-English test set.
\end{itemize}

This paper is structured as follows: in Section~\ref{sect:related} we overview previous work upon which we build; Section~\ref{sect:content} details the model and data settings, and the released test set; and Section~\ref{sect:benchmarks} compares our model to other public models and to state-of-the-art results in academic competitions.

The model can be downloaded at 
\url{https://github.com/naver/covid19-nmt}: the repository consists in a model checkpoint that is compatible with \textit{Fairseq}~\citep{fairseq}, and a script to preprocess the input text.

\section{Related Work}
\label{sect:related}
In order to serve its purpose, our model should be able to process multilingual input sentences, and generate tailored translations for COVID-19-related sentences.
As far as NMT models are concerned, both multilingual and domain-specific sentences are just sequences of plain tokens that should be distinguished internally and handled in a separate manner depending on the multiple languages or domains.
Due to this commonality in both fields of MNMT and domain adaptation of NMT models, they can be broadly categorized into two groups: 1) data-centric and 2) model-centric~\citep{chu-wang-2018-survey}.

The former focuses on the preparation of the training data such as handling and selecting from multi-domain~\citep{kobus2016,tars2018multi} or multilingual parallel corpora~\citep{aharoni2019massively,tan2019multilingual}; and generating synthetic parallel data from monolingual corpora~\citep{sennrichHB15a,edunov-etal-2018-understanding}.

The model-centric approaches, on the other hand, center on adjusting the training objectives~\citep{wang-etal-2017-instance,tan2018multilingual}; modifying the model architectures~\citep{vazquez2018multilingual,dou2019unsupervised}; and tweaking the decoding procedure~\citep{hasler2018neural,dou2019domain}.

While the two types of approaches are orthogonal and can be utilized in tandem, our released model is trained using data-centric approaches.
One of the frequently used data-centric methods for handling sentences of multiple languages and domains is simply prepending a special token that indicates the target language or domain that the sentence should be translated into~\citep{kobus2016,aharoni2019massively}.
By feeding the task-specific meta-information via the reserved tags, we signal the model to treat the following input tokens accordingly.
Recent works show that this method is also applicable to generating diverse translations~\citep{shu-etal-2019-generating} and translations in specific styles~\citep{madaan2020politeness}.
% AB: we don't use language tags

In addition, back-translation of target monolingual or domain-specific sentences is often conducted in order to augment the low-resource data~\citep{edunov-etal-2018-understanding,Hu_2019}.
Afterward, the back-translated data (and existing parallel data) can be filtered~\citep{xu2019improving} or treated with varying amount of importance~\citep{wang2019improving} using data selection methods. Back-translated sentences can be tagged to achieve even better results \citep{caswell-etal-2019-tagged}.

\smallskip

While myriads of research works on MNMT and domain adaptation exist, the number of publicly available pre-trained NMT models is still low.
For example, \texttt{Fairseq}, a popular sequence-to-sequence toolkit maintained by Facebook AI Research, has released ten uni-directional models for translating English, French, German, and Russian sentences.\footnote{\url{https://github.com/pytorch/fairseq/blob/master/examples/translation/README.md}}
For its widespread usage, we trained our model using this toolkit.

A large number of public MT models are available thanks to OPUS-MT,\footnote{\url{https://github.com/Helsinki-NLP/OPUS-MT-train}} created by the Helsinki-NLP group.
Utilizing the OPUS corpora~\citep{tiedemann-2012-parallel}, more than a thousand MT models are trained and released, including several multilingual models which we use to compare with our model.

To the best of our knowledge, we release the first public MNMT model that is capable of producing tailored translations for the biomedical domain.

\smallskip

The COVID-19 pandemic has shown the need for multilingual access to hygiene and safety guidelines and policies~\citep{wired2020}.
As an example of crowd-sourced translation, we point out ``The COVID Translate Project''\footnote{\url{https://covidtranslate.org/}} which allowed the translation of 75 pages of guidelines for public agents and healthcare workers, from Korean into English in a matter of days.
Although our model could assist in furthering such initiatives, we do not recommend relying solely on our model for translating such guidelines, where quality is of uttermost importance.
However, the huge amount of digital content created in the last months around the pandemic makes such professional translations of all that content not only infeasible, but sometimes unnecessary depending on the objective.
For instance, we believe that the release of this model can unlock the possibility of large-scale translation with the aim of conducting data analysis on the reaction of the media and society on the matter.

\section{Model Settings and Training Data}
\label{sect:content}

The model uses a variant of the Transformer Big architecture \citep{vaswani2017} with a shallower decoder: 16 attention heads, 6 encoder layers, 3 decoder layers, an embedding size of 1024, and a feed-forward dimension of 8192 in the encoder and 4096 in the decoder.

As all language pairs have English as their target language, no special token for target language was used (language detection can be performed internally by the model). % AB: we trained a first version with language codes but saw no difference in BLEU

As the model performs many-to-English translation, its encoder should be able to hold most of the complexity.
Thus, we increase the capacity of the encoder by doubling the default size of the feed-forward layer as in~\cite{wmt19fair}.
%\todo{ZM: I added the reference, please check if it is correct.}

On the other hand, previous works~\citep{clinchant-etal-2019-use, kasai2020deep} have shown that it is possible to reduce the number of decoder layers without sacrificing much performance, allowing both faster inference, and smaller network size.

During training, regularization was done with a dropout of $0.1$ and label smoothing of $0.1$.
For optimization, we used Adam~\citep{kingma2014adam} with warm-up, and maximum learning rate of $0.001$.
The model was trained for 10 epochs and the best checkpoint was selected based on perplexity on the validation set.

\smallskip

As training data, we used the standard open-accessible datasets, including biomedical data whenever available, for example, the ``Corona Crisis Corpora''~\citep{CCCTAUS}.
Following our past success in domain adaptation~\citep{berard2019}, we used domain tokens~\citep{kobus2016} to differentiate between domains, allowing multi-domain translation with a single model.
We initially experimented with more tags, and combinations of tags (e.g., \texttt{medical $\rightarrow$ patent} or \texttt{medical $\rightarrow$ political}) to allow for more fine-grained control of the resulting translation.
The results however were not very conclusive, and often under-performed.
An exception worth noting was the case of transcribed data such as \textit{TED} talks, and \textit{OpenSubtitles}, which are not the main targets of this work. Therefore, for simplicity, we used only two tags: \texttt{medical} and \texttt{back-translation}. No tag was used with training data that does not belong to one of these two categories.

In addition to biomedical data, we also used back-translated data, although only for Korean, the language with the smallest amount of training data (13.8M sentences). Like \citet{arivazhagan2019massively}, we used a temperature parameter of 5, to give more chance to Korean. Additionally, the biomedical data was oversampled by a factor of 2.
Table~\ref{tab:data} details the amount of training sentences used for each language and each domain tag.

\bigskip

As for pre-processing, we cleaned the available data by conducting white-space normalization and NFKC normalization. We filtered noisy sentence pairs based on length (min. 1 token, max. 200), and automatic language identification with \texttt{langid.py} \citep{lui-baldwin-2012-langid}. %\footnote{\url{https://github.com/saffsd/langid.py}}

We trained a lower-cased shared BPE model using \textit{SentencePiece}~\citep{kudo2018sentencepiece} by using 6M random lines for each language (including English). We filtered out single characters with fewer than 20 occurrences from the vocabulary. This results in a shared vocabulary of size 76k.

We reduced the English vocabulary size to speed up training and inference, by setting a BPE frequency threshold of 20, which gives a target vocabulary of size 38k. To get the benefits of a shared vocabulary (i.e., tied source/target embeddings) we sorted the source Fairseq dictionary to put the 38k English tokens at the beginning, which lets us easily share the embedding matrix between the encoder and the decoder.\footnote{We modified the released checkpoint for it to work out-of-the box with vanilla Fairseq.}

The BPE segmentation is followed by inline-casing \citep{berard2019}, where each token is lower-cased and directly followed by a special token specifying its case (\texttt{<T>} for title case, \texttt{<U>} for all caps, no token for lower-case). Word-pieces whose original case is undefined (e.g., ``MacDonalds'') are split again into word-pieces with defined case (``mac'' and ``donalds'').

\begin{table}[t]
    \centering
    \centering
\begin{tabular}{|l|r|rrr|}
\hline
Language & Total & General & BT & Biomed. \\
\hline
French   & 128.8 & 125.0   & --                 & 3.8        \\
Spanish  & 92.9  & 90.8    & --                 & 2.1        \\
German   & 87.3  & 84.8    & --                 & 2.5        \\
Italian  & 45.6  & 44.9    & --                 & 0.7        \\
Korean   & 13.8  & 5.7     & 8.0              & 0.1        \\
\hline
Total      & 368.4 & 351.2   & 8.0              & 9.2        \\
\hline
\end{tabular}
    \caption{Amount of training data in millions of sentences. BT refers to back-translation from English to Korean.}
    \label{tab:data}
\end{table}

\subsection{New Korean-English Test Set}
To benchmark the performance on the COVID-19 domain, we built an in-domain test set for Korean-English, as it is the only language pair that is not included in the Corona Crisis Corpora.

The test set contains 758 Korean-English sentence pairs, obtained by having English sentences translated into Korean by four professional Korean translators.
We note that any acronym written without its full form in the source sentence is kept the same in the translation unless it is very widely used in general.
The English sentences were distributed among the translators with the same guidelines to get consistent tone and manner.

We gathered English sentences from two sources: 1) The official English guidelines and reports from Korea Centers for Disease Control and Prevention (KCDC)\footnote{\url{http://ncov.mohw.go.kr/en}} under Ministry of Health and Welfare of South Korea (258 sentences); and 2) Abstracts of biomedical papers on SARS-CoV-2 and COVID-19 from \textit{arXiv},\footnote{\url{https://arxiv.org}} \textit{medRxiv}\footnote{\url{https://www.medrxiv.org}} and \textit{bioRxiv}\footnote{\url{https://www.biorxiv.org}} (500 sentences).
The sentences were handpicked, focusing on covering diverse aspects of the pandemic, including safety guidelines, government briefings, clinical tests, and biomedical experimentation.
%\todo{ZM: I've asked for more details on the translation process and QA. Will update the sect. if needed.}

\section{Benchmarks}
\label{sect:benchmarks}

We benchmarked the released multilingual models against: 1) reported numbers in the literature, and 2) other publicly released models.
We used OPUS-MT, a large collection (1000+) of pre-trained models released by the NLP group at University of Helsinki. Note that these models were trained with much smaller amounts of training data.

We note that the biomedical test sets (\textit{Medline}) are very small (around 600 lines). %, and we computed the scores for translations against the untokenized references, which might differ from the official task (from which we reported the numbers).
We do not report comparison for Spanish-English \textit{newstest2013}, as the latest reported numbers are outdated (the best WMT entry achieved $30.4$).
%Spanish, Italian and Korean were added later to the IWSLT challenge, and were not part of the competition at that time.

Our single model obtains competitive results on ``generic'' test sets (\textit{News} and \textit{IWSLT}), on par with the state of the art. We also obtain strong results on the biomedical test sets. Note the SOTA models were trained to maximize performance in the very specific Medline domain, for which training data is provided. While we included this data in our tagged biomedical data, we did not fine-tune aggressively over it.

Table~\ref{tab:benchmark_ko_en} shows the BLEU scores for the Korean-English COVID-19 test set.
The results greatly outperform existing public Korean-English models, even more so than with the IWSLT test sets (Table~\ref{tab:benchmark}).
%Using the \texttt{medical} tag provides a small but consistent boost in BLEU score.

% \todo{ZM: Talk more about the results.}
% \texttt{BLEU for COVID is higher than for IWSLT, even without the medical tag. Is that because of the back-translated biomedical dataset? Also possibly because of the fact that the input English sentences are translationese.}

% Qualitatively, we observe that the translations generated with the \texttt{medical} tag tend to .. \texttt{Ran compare-mt to observe the differences on KO-EN, but not much diff. there (because of the BT effect?). Perhaps try on other lang-pairs?}

\begin{table}[t]
    \center
    \begin{tabular}{|l|ccc|}
\hline
Model                                      & arXiv & KCDC & Full  \\ \hline
Ours                                        & 36.5  & 38.3 & 37.2 \\
Ours (\texttt{medical}) & 36.6  & 38.6 & 37.4 \\
% VN: I updated numbers with fixed problem in helsinki preprocessing
OPUS-MT               & 18.7 & 19.0 &  18.8    \\ \hline
\end{tabular}
    \caption{Benchmark of the released model on new Korean-English COVID-19 test set.}
    \label{tab:benchmark_ko_en}
\end{table}

\begin{table}[t]
    \centering
    \footnotesize
\begin{tabular}{|ll|lll|}
\hline
Language                 & Model                  & News & Medline & IWSLT \\
\hline
\multirow{3}{*}{French}  & Ours               & \textbf{41.00}       & \textbf{36.16}            & \textbf{41.09}      \\
                         & SOTA              & 40.22\textsuperscript{\ref{fn:nleugc}} & 35.56\textsuperscript{\ref{fn:biomedical19}}            & -- \\% \textbf{41.70}\textsuperscript{\ref{fn:nleugc}}      \\
                         %& OPUS-MT   &    ~~33.30~~      &  29.10                &                  & 36.40           \\
                         & OPUS-MT   &    36.80      &   33.60           & 38.90           \\
\hline
\multirow{3}{*}{German}  & Ours               & \textbf{41.28}    & \textbf{29.76}             & 31.55      \\
                         & SOTA              & 40.98\textsuperscript{\ref{fn:fairnews}}    & 28.82\textsuperscript{\ref{fn:biomedical19}}            & \textbf{32.01}\textsuperscript{\ref{fn:fairnews}}      \\
                         & OPUS-MT  &  39.50        &  
                         28.10           & 30.30           \\  
\hline
\multirow{3}{*}{Spanish} & Ours               & \textbf{36.63}    & \textbf{46.18}            & \textbf{48.79}      \\
                         & SOTA              & --       & 43.03\textsuperscript{\ref{fn:biomedical19}}            & --         \\
                         & OPUS-MT &   30.30       &     43.30              &46.10            \\
\hline
\multirow{3}{*}{Italian} & Ours               & \cellcolor{gray!15}       & \cellcolor{gray!15}               & \textbf{42.18}      \\
                         & SOTA              & \cellcolor{gray!15}       & \cellcolor{gray!15}                 & --            \\
                         & OPUS-MT  &   \cellcolor{gray!15}   %32.40   
                         &      \cellcolor{gray!15}              & 39.70\\
\hline
\multirow{3}{*}{Korean}  & Ours               & \cellcolor{gray!15}       & \cellcolor{gray!15}                & \textbf{21.33}      \\
                         & SOTA              & \cellcolor{gray!15}         &\cellcolor{gray!15}                  & --           \\
                         & OPUS-MT  &\cellcolor{gray!15}          &\cellcolor{gray!15}                     &  17.60    \\
\hline
\end{tabular}

    \caption{Benchmark of the released model against the best reported numbers and the public OPUS-MT models.
    Our scores on the medical test sets are obtained with the \texttt{medical} tag. No tag is used with the other test sets.
    The SOTA numbers for \emph{News} and \emph{IWSLT} were either obtained by running the corresponding models~\citep{berard2019,wmt19fair}. For \emph{Medline}, we copied the results from the WMT19 Biomedical Task report~\citep{wmt19biomedical}.
    \textit{News} consists in the \emph{newstest2019} for German (WMT19 News Task test set), \emph{newstest2014} for French, and \emph{newstest2013} for Spanish. 
    For \textit{IWSLT}, the 2017 test set is used for all but Spanish, where the 2016 one is used.}
    \label{tab:benchmark}
\end{table}

\renewcommand*{\thefootnote}{\fnsymbol{footnote}}
\setcounter{footnote}{0}
\stepcounter{footnote}
\footnotetext{NLE @ WMT19 Robustness Task~\citep{berard2019}\label{fn:nleugc}}  
\stepcounter{footnote}
\footnotetext{FAIR @ WMT19 News Task~\citep{wmt19fair}\label{fn:fairnews}}  
\stepcounter{footnote}
%\footnotetext{Reported results at WMT18-biomedical~\citep{wmt18biomedical}\label{fn:biomedical18}}  
%\stepcounter{footnote}
\footnotetext{Reported results in the WMT19 Biomedical Task~\citep{wmt19biomedical}\label{fn:biomedical19}}

\section{Conclusion}
We describe the release of a multilingual translation model that supports translation in both the general and biomedical domains.
Our model is trained on more than 350M sentences, covering French, Spanish, German, Italian and Korean (into English).
Benchmarks on public test sets show its strength across domains. In particular, we evaluated the model in the biomedical domain, where it performs near state-of-the-art, but with the advantage of being a single model that operates on several languages.
To address the shortage of Korean-English data, we also release a dataset of 758 sentence pairs covering recent biomedical text about COVID-19.

Our aim is to support research studying the international impact that this crisis is causing, at a societal, economical and healthcare level. 

%\todo{run: NEWSTEST.it.en for our model, our model and Helsinki [DONE] on new test data}
\newpage

\bibliography{acl2020}

\begin{thebibliography}{36}
\expandafter\ifx\csname natexlab\endcsname\relax\def\natexlab#1{#1}\fi

\bibitem[{Aharoni et~al.(2019)Aharoni, Johnson, and
  Firat}]{aharoni2019massively}
Roee Aharoni, Melvin Johnson, and Orhan Firat. 2019.
\newblock Massively multilingual neural machine translation.
\newblock \emph{arXiv preprint arXiv:1903.00089}.

\bibitem[{Anastasopoulos and Neubig(2019)}]{anastasopoulos2019should}
Antonios Anastasopoulos and Graham Neubig. 2019.
\newblock Should all cross-lingual embeddings speak english?
\newblock \emph{arXiv preprint arXiv:1911.03058}.

\bibitem[{Arivazhagan et~al.(2019)Arivazhagan, Bapna, Firat, Lepikhin, Johnson,
  Krikun, Chen, Cao, Foster, Cherry et~al.}]{arivazhagan2019massively}
Naveen Arivazhagan, Ankur Bapna, Orhan Firat, Dmitry Lepikhin, Melvin Johnson,
  Maxim Krikun, Mia~Xu Chen, Yuan Cao, George Foster, Colin Cherry, et~al.
  2019.
\newblock Massively multilingual neural machine translation in the wild:
  Findings and challenges.
\newblock \emph{arXiv preprint arXiv:1907.05019}.

\bibitem[{Banda et~al.(2020)Banda, Tekumalla, Wang, Yu, Liu, Ding, and
  Chowell}]{banda2020}
Juan~M Banda, Ramya Tekumalla, Guanyu Wang, Jingyuan Yu, Tuo Liu, Yuning Ding,
  and Gerardo Chowell. 2020.
\newblock A large-scale covid-19 twitter chatter dataset for open scientific
  research--an international collaboration.
\newblock \emph{arXiv preprint arXiv:2004.03688}.

\bibitem[{Bawden et~al.(2019)Bawden, Bretonnel~Cohen, Grozea, Jimeno~Yepes,
  Kittner, Krallinger, Mah, Neveol, Neves, Soares, Siu, Verspoor, and
  Vicente~Navarro}]{wmt19biomedical}
Rachel Bawden, Kevin Bretonnel~Cohen, Cristian Grozea, Antonio Jimeno~Yepes,
  Madeleine Kittner, Martin Krallinger, Nancy Mah, Aurelie Neveol, Mariana
  Neves, Felipe Soares, Amy Siu, Karin Verspoor, and Maika Vicente~Navarro.
  2019.
\newblock \href {https://doi.org/10.18653/v1/W19-5403} {Findings of the {WMT}
  2019 biomedical translation shared task: Evaluation for {MEDLINE} abstracts
  and biomedical terminologies}.
\newblock In \emph{Proceedings of the Fourth Conference on Machine Translation
  (Volume 3: Shared Task Papers, Day 2)}, pages 29--53, Florence, Italy.
  Association for Computational Linguistics.

\bibitem[{Berard et~al.(2019)Berard, Calapodescu, and Roux}]{berard2019}
Alexandre Berard, Ioan Calapodescu, and Claude Roux. 2019.
\newblock \href {https://doi.org/10.18653/v1/W19-5361} {Naver labs {E}urope{'}s
  systems for the {WMT}19 machine translation robustness task}.
\newblock In \emph{Proceedings of the Fourth Conference on Machine Translation
  (Volume 2: Shared Task Papers, Day 1)}, pages 526--532, Florence, Italy.
  Association for Computational Linguistics.

\bibitem[{Caswell et~al.(2019)Caswell, Chelba, and
  Grangier}]{caswell-etal-2019-tagged}
Isaac Caswell, Ciprian Chelba, and David Grangier. 2019.
\newblock \href {https://doi.org/10.18653/v1/W19-5206} {Tagged
  back-translation}.
\newblock In \emph{Proceedings of the Fourth Conference on Machine Translation
  (Volume 1: Research Papers)}, pages 53--63, Florence, Italy. Association for
  Computational Linguistics.

\bibitem[{Chu and Wang(2018)}]{chu-wang-2018-survey}
Chenhui Chu and Rui Wang. 2018.
\newblock A survey of domain adaptation for neural machine translation.
\newblock In \emph{Proceedings of the 27th International Conference on
  Computational Linguistics}, pages 1304--1319, Santa Fe, New Mexico, USA.
  Association for Computational Linguistics.

\bibitem[{Clinchant et~al.(2019)Clinchant, Jung, and
  Nikoulina}]{clinchant-etal-2019-use}
Stephane Clinchant, Kweon~Woo Jung, and Vassilina Nikoulina. 2019.
\newblock \href {https://doi.org/10.18653/v1/D19-5611} {On the use of {BERT}
  for neural machine translation}.
\newblock In \emph{Proceedings of the 3rd Workshop on Neural Generation and
  Translation}, pages 108--117, Hong Kong. Association for Computational
  Linguistics.

\bibitem[{Croquet(2020)}]{lemonde2020}
Pauline Croquet. 2020.
\newblock \href
  {https://www.lemonde.fr/pixels/article/2020/05/15/a-la-bnf-les-archivistes-du-web-sauvegardent-l-internet-francais-du-confinement_6039704_4408996.html}
  {Comment les archivistes de la {BNF} sauvegardent la mémoire du confinement
  sur internet}.
\newblock Accessed June 2020.

\bibitem[{Dou et~al.(2019{\natexlab{a}})Dou, Hu, Anastasopoulos, and
  Neubig}]{dou2019unsupervised}
Zi-Yi Dou, Junjie Hu, Antonios Anastasopoulos, and Graham Neubig.
  2019{\natexlab{a}}.
\newblock Unsupervised domain adaptation for neural machine translation with
  domain-aware feature embeddings.
\newblock In \emph{Proceedings of the 2019 Conference on Empirical Methods in
  Natural Language Processing and the 9th International Joint Conference on
  Natural Language Processing (EMNLP-IJCNLP)}, pages 1417--1422.

\bibitem[{Dou et~al.(2019{\natexlab{b}})Dou, Wang, Hu, and
  Neubig}]{dou2019domain}
Zi-Yi Dou, Xinyi Wang, Junjie Hu, and Graham Neubig. 2019{\natexlab{b}}.
\newblock Domain differential adaptation for neural machine translation.
\newblock \emph{arXiv preprint arXiv:1910.02555}.

\bibitem[{Edunov et~al.(2018)Edunov, Ott, Auli, and
  Grangier}]{edunov-etal-2018-understanding}
Sergey Edunov, Myle Ott, Michael Auli, and David Grangier. 2018.
\newblock \href {https://doi.org/10.18653/v1/D18-1045} {Understanding
  back-translation at scale}.
\newblock In \emph{Proceedings of the 2018 Conference on Empirical Methods in
  Natural Language Processing}, pages 489--500, Brussels, Belgium. Association
  for Computational Linguistics.

\bibitem[{Hasler et~al.(2018)Hasler, de~Gispert, Iglesias, and
  Byrne}]{hasler2018neural}
Eva Hasler, Adri{\`a} de~Gispert, Gonzalo Iglesias, and Bill Byrne. 2018.
\newblock \href {https://doi.org/10.18653/v1/N18-2081} {Neural machine
  translation decoding with terminology constraints}.
\newblock In \emph{Proceedings of the 2018 Conference of the North {A}merican
  Chapter of the Association for Computational Linguistics: Human Language
  Technologies, Volume 2 (Short Papers)}, pages 506--512, New Orleans,
  Louisiana. Association for Computational Linguistics.

\bibitem[{Hu et~al.(2019)Hu, Xia, Neubig, and Carbonell}]{Hu_2019}
Junjie Hu, Mengzhou Xia, Graham Neubig, and Jaime Carbonell. 2019.
\newblock \href {https://doi.org/10.18653/v1/p19-1286} {Domain adaptation of
  neural machine translation by lexicon induction}.
\newblock \emph{Proceedings of the 57th Annual Meeting of the Association for
  Computational Linguistics}.

\bibitem[{Kasai et~al.(2020)Kasai, Pappas, Peng, Cross, and
  Smith}]{kasai2020deep}
Jungo Kasai, Nikolaos Pappas, Hao Peng, James Cross, and Noah~A. Smith. 2020.
\newblock \href {http://arxiv.org/abs/2006.10369} {Deep encoder, shallow
  decoder: Reevaluating the speed-quality tradeoff in machine translation}.

\bibitem[{Kingma and Ba(2014)}]{kingma2014adam}
Diederik~P Kingma and Jimmy Ba. 2014.
\newblock Adam: A method for stochastic optimization.
\newblock \emph{arXiv preprint arXiv:1412.6980}.

\bibitem[{Kobus et~al.(2016)Kobus, Crego, and Senellart}]{kobus2016}
Catherine Kobus, Josep Crego, and Jean Senellart. 2016.
\newblock Domain control for neural machine translation.
\newblock \emph{arXiv preprint arXiv:1612.06140}.

\bibitem[{Kudo and Richardson(2018)}]{kudo2018sentencepiece}
Taku Kudo and John Richardson. 2018.
\newblock Sentencepiece: A simple and language independent subword tokenizer
  and detokenizer for neural text processing.
\newblock \emph{arXiv preprint arXiv:1808.06226}.

\bibitem[{Lui and Baldwin(2012)}]{lui-baldwin-2012-langid}
Marco Lui and Timothy Baldwin. 2012.
\newblock langid.py: An off-the-shelf language identification tool.
\newblock In \emph{Proceedings of the {ACL} 2012 System Demonstrations}, pages
  25--30, Jeju Island, Korea. Association for Computational Linguistics.

\bibitem[{Madaan et~al.(2020)Madaan, Setlur, Parekh, Poczos, Neubig, Yang,
  Salakhutdinov, Black, and Prabhumoye}]{madaan2020politeness}
Aman Madaan, Amrith Setlur, Tanmay Parekh, Barnabas Poczos, Graham Neubig,
  Yiming Yang, Ruslan Salakhutdinov, Alan~W Black, and Shrimai Prabhumoye.
  2020.
\newblock \href {http://arxiv.org/abs/2004.14257} {Politeness transfer: A tag
  and generate approach}.

\bibitem[{Mc{C}ulloch(2020)}]{wired2020}
Gretchen Mc{C}ulloch. 2020.
\newblock Covid-19 is history’s biggest translation challenge.

\bibitem[{Ng et~al.(2019)Ng, Yee, Baevski, Ott, Auli, and Edunov}]{wmt19fair}
Nathan Ng, Kyra Yee, Alexei Baevski, Myle Ott, Michael Auli, and Sergey Edunov.
  2019.
\newblock \href {https://doi.org/10.18653/v1/W19-5333} {{F}acebook {FAIR}{'}s
  {WMT}19 news translation task submission}.
\newblock In \emph{Proceedings of the Fourth Conference on Machine Translation
  (Volume 2: Shared Task Papers, Day 1)}, pages 314--319, Florence, Italy.
  Association for Computational Linguistics.

\bibitem[{Ott et~al.(2019)Ott, Edunov, Baevski, Fan, Gross, Ng, Grangier, and
  Auli}]{fairseq}
Myle Ott, Sergey Edunov, Alexei Baevski, Angela Fan, Sam Gross, Nathan Ng,
  David Grangier, and Michael Auli. 2019.
\newblock fairseq: A fast, extensible toolkit for sequence modeling.
\newblock \emph{arXiv preprint arXiv:1904.01038}.

\bibitem[{Sennrich et~al.(2015)Sennrich, Haddow, and Birch}]{sennrichHB15a}
Rico Sennrich, Barry Haddow, and Alexandra Birch. 2015.
\newblock \href {http://arxiv.org/abs/1511.06709} {Improving neural machine
  translation models with monolingual data}.
\newblock \emph{CoRR}, abs/1511.06709.

\bibitem[{Shu et~al.(2019)Shu, Nakayama, and Cho}]{shu-etal-2019-generating}
Raphael Shu, Hideki Nakayama, and Kyunghyun Cho. 2019.
\newblock \href {https://doi.org/10.18653/v1/P19-1177} {Generating diverse
  translations with sentence codes}.
\newblock In \emph{Proceedings of the 57th Annual Meeting of the Association
  for Computational Linguistics}, pages 1823--1827, Florence, Italy.
  Association for Computational Linguistics.

\bibitem[{Tan et~al.(2019{\natexlab{a}})Tan, Chen, He, Xia, Qin, and
  Liu}]{tan2019multilingual}
Xu~Tan, Jiale Chen, Di~He, Yingce Xia, Tao Qin, and Tie-Yan Liu.
  2019{\natexlab{a}}.
\newblock Multilingual neural machine translation with language clustering.
\newblock \emph{arXiv preprint arXiv:1908.09324}.

\bibitem[{Tan et~al.(2019{\natexlab{b}})Tan, Ren, He, Qin, and
  Liu}]{tan2018multilingual}
Xu~Tan, Yi~Ren, Di~He, Tao Qin, and Tie-Yan Liu. 2019{\natexlab{b}}.
\newblock Multilingual neural machine translation with knowledge distillation.
\newblock In \emph{International Conference on Learning Representations}.

\bibitem[{Tars and Fishel(2018)}]{tars2018multi}
Sander Tars and Mark Fishel. 2018.
\newblock Multi-domain neural machine translation.
\newblock \emph{arXiv preprint arXiv:1805.02282}.

\bibitem[{TAUS(2020)}]{CCCTAUS}
TAUS. 2020.
\newblock \href {https://md.taus.net/corona} {Taus corona crisis corpus}.

\bibitem[{Tiedemann(2012)}]{tiedemann-2012-parallel}
J{\"o}rg Tiedemann. 2012.
\newblock Parallel data, tools and interfaces in {OPUS}.
\newblock In \emph{Proceedings of the Eighth International Conference on
  Language Resources and Evaluation ({LREC}'12)}, pages 2214--2218, Istanbul,
  Turkey. European Language Resources Association (ELRA).

\bibitem[{Vaswani et~al.(2017)Vaswani, Shazeer, Parmar, Uszkoreit, Jones,
  Gomez, Kaiser, and Polosukhin}]{vaswani2017}
Ashish Vaswani, Noam Shazeer, Niki Parmar, Jakob Uszkoreit, Llion Jones,
  Aidan~N Gomez, {\L}ukasz Kaiser, and Illia Polosukhin. 2017.
\newblock Attention is all you need.
\newblock In \emph{Advances in neural information processing systems}, pages
  5998--6008.

\bibitem[{V{\'a}zquez et~al.(2018)V{\'a}zquez, Raganato, Tiedemann, and
  Creutz}]{vazquez2018multilingual}
Ra{\'u}l V{\'a}zquez, Alessandro Raganato, J{\"o}rg Tiedemann, and Mathias
  Creutz. 2018.
\newblock Multilingual nmt with a language-independent attention bridge.
\newblock \emph{arXiv preprint arXiv:1811.00498}.

\bibitem[{Wang et~al.(2017)Wang, Utiyama, Liu, Chen, and
  Sumita}]{wang-etal-2017-instance}
Rui Wang, Masao Utiyama, Lemao Liu, Kehai Chen, and Eiichiro Sumita. 2017.
\newblock \href {https://doi.org/10.18653/v1/D17-1155} {Instance weighting for
  neural machine translation domain adaptation}.
\newblock In \emph{Proceedings of the 2017 Conference on Empirical Methods in
  Natural Language Processing}, pages 1482--1488, Copenhagen, Denmark.
  Association for Computational Linguistics.

\bibitem[{Wang et~al.(2019)Wang, Liu, Wang, Luan, and Sun}]{wang2019improving}
Shuo Wang, Yang Liu, Chao Wang, Huanbo Luan, and Maosong Sun. 2019.
\newblock Improving back-translation with uncertainty-based confidence
  estimation.
\newblock \emph{arXiv preprint arXiv:1909.00157}.

\bibitem[{Xu et~al.(2019)Xu, Ko, and Seo}]{xu2019improving}
Guanghao Xu, Youngjoong Ko, and Jungyun Seo. 2019.
\newblock Improving neural machine translation by filtering synthetic parallel
  data.
\newblock \emph{Entropy}, 21(12):1213.

\end{thebibliography}
\bibliographystyle{acl_natbib}

%\appendix

%\section{Appendices}
%\label{sec:appendix}

\end{document}